\documentclass[11pt,a4paper]{article}
\usepackage[hyperref]{acl2020}
\usepackage{times}
\usepackage{latexsym}

\usepackage{graphicx}
\usepackage{amsmath}
\usepackage{microtype}

\aclfinalcopy 

\title{Robust Prediction of Punctuation and Truecasing \\ for Medical ASR}

\author{Monica Sunkara \quad Srikanth Ronanki \quad Kalpit Dixit \quad Sravan Bodapati \quad Katrin Kirchhoff\\
  Amazon AWS AI, USA\\
  {\tt \{sunkaral, ronanks\}@amazon.com} \\
}

\begin{document}

\maketitle


\begin{abstract}


Automatic speech recognition (ASR) systems in the medical domain that focus on transcribing clinical dictations and doctor-patient conversations often pose many challenges due to the complexity of the domain. ASR output typically undergoes automatic punctuation to enable users to speak naturally, without having to vocalise awkward and explicit punctuation commands, such  as ``period'', ``add  comma'' or ``exclamation point'', while truecasing enhances user readability and improves the performance of downstream NLP tasks. This paper proposes a conditional joint modeling framework for prediction of punctuation and truecasing using pretrained masked language models such as BERT, BioBERT and RoBERTa. We also present techniques for domain and task specific adaptation by fine-tuning masked language models with medical domain data.  Finally, we improve the robustness of the model against common errors made in ASR by performing data augmentation. Experiments performed on dictation and conversational style corpora show that our proposed model achieves $\sim$5\% absolute improvement on ground truth text and $\sim$10\% improvement on ASR outputs over baseline models under F1 metric.
  
\end{abstract}

\section{Introduction}
Medical ASR systems automatically transcribe medical speech found in a variety of use cases like physician-dictated notes \cite{DictASR}, telemedicine and even doctor-patient conversations \cite{ConvASR}, without any human intervention. These systems ease the burden of long hours of administrative work and also promote better engagement with patients. However, the generated ASR outputs are typically devoid of punctuation and truecasing thereby making it difficult to comprehend. Furthermore, their recovery improves the accuracy of subsequent natural language understanding algorithms \cite{nlp11, nlp22} to identify information such as patient diagnosis, treatments, dosages, symptoms and signs. Typically, clinicians explicitly dictate the punctuation commands like ``period'', ``add comma'' etc., and a postprocessing component takes care of punctuation restoration. This process is usually error-prone as the clinicians may struggle with appropriate punctuation insertion during dictation. Moreover, doctor-patient conversations lack explicit vocalization of punctuation marks motivating the need for automatic prediction of punctuation and truecasing. In this work, we aim to solve the problem of automatic punctuation and truecasing restoration to medical ASR system text outputs.

Most recent approaches to punctuation and truecasing restoration problem rely on deep learning \cite{chunk, salloum2017deep}. Although it is a well explored problem in the literature, most of these improvements do not directly translate to great real world performance in all settings. For example, unlike general text, it is a harder problem to solve when applied to the medical domain for various reasons and we illustrate each of them:
\begin{itemize}
\item \textbf{Large vocabulary:} ASR systems in the medical domain have a large set of domain-specific vocabulary and several abbreviations. Owing to the domain specific data set and the open vocabulary in  LVCSR (large-vocabulary continuous speech recognition) outputs, we often run into OOV (out of vocabulary) or rare word problems. Furthermore, a large vocabulary set leads to data sparsity issues. We address both these problems by using subword models. Subwords have been shown to work well in open-vocabulary speech recognition and several NLP tasks \cite{subwordnmt, bodapati-etal-2019-neural}. We compare word and subword models across different architectures and show that subword models consistently outperform the former.
\item \textbf{Data scarcity:} Data scarcity is one of the major bottlenecks in supervised learning. When it comes to the medical domain, obtaining data is not as straight-forward as some of the other domains where abundance of text is available. 
On the other hand, obtaining large amounts of data is a tedious and costly process; procuring and maintaining it could be a challenge owing to the strict privacy laws. We overcome the data scarcity problem, by using pretrained masked language models like BERT \cite{bert} and its successors \cite{roberta, xlnet} which have successfully been shown to produce state-of-the-art results when finetuned for several downstream tasks like question answering and language inference. We approach the prediction task as a sequence labeling problem and jointly learn punctuation and truecasing. We show that finetuning a pretrained model with a very small medical dataset ($\sim$500k\ words) has $\sim$5\% absolute performance improvement in terms of F1 compared to a model trained from scratch. We further boost the performance by first finetuning the masked language model to the medical speech domain and then to the downstream task.
\item \textbf{ASR Robustness:} Models trained on ground truth data are not exposed to typical errors in speech recognition and perform poorly when evaluated on ASR outputs. Our objective is to make the punctuation prediction and truecasing more robust to speech recognition errors and establish a mechanism to test the performance of the model quantitatively. To address this issue, we propose a data augmentation based approach using n-best lists from ASR. 
\end{itemize}



The contributions of this work are: 
\begin{itemize} 
\item A general post-processing framework for conditional joint labeling of punctuation and truecasing for medical ASR (clinical dictation and conversations). 
\item An analysis comparing different embeddings that are suitable for the medical domain. An in-depth analysis of the effectiveness of using pretrained masked language models like BERT and its successors to address the data scarcity problem.
\item Techniques for effective domain and task adaptation using Masked Language Model (\textit{MLM}) finetuning of BERT on medical domain data to boost the downstream task performance.
\item Method for enhancing robustness of the  models via data augmentation with n-best lists (from ASR output) to the ground truth during training to improve performance on ASR hypothesis at inference time.
\end{itemize}

The rest of this paper is organized as follows. Section \ref{sec:related_work} presents related work on punctuation and truecasing restoration. Section \ref{sec:proposed_framework} introduces the model architecture used in this paper and describes various techniques for improving accuracy and robustness. The experimental evaluation and results are discussed in Section \ref{sec:experiments} and finally, Section \ref{sec:conclusion} presents the conclusions.

\section{Related work}
\label{sec:related_work}
Several researchers have proposed a number of methodologies such as the use of probabilistic machine learning models, neural network models, and the acoustic fusion approaches for punctuation prediction. We review related work in these areas below. 

\subsection{Earlier methods}
\label{ssec:earlier_methods}
In earlier efforts, punctuation prediction has been approached by using finite state or hidden Markov models \cite{fst1,fst2}. Several other approaches addressed it as a language modeling problem by predicting the most probable sequence of words with punctuation marks inserted \cite{lm2, lm3, lm5}. Some others used conditional random fields (CRFs) \cite{CRF2, CRF1} and maximum entropy using n-grams \cite{mem1}. The rise of stronger machine learning techniques such as deep and/or recurrent neural networks replaced these conventional models.

\subsection{ Using acoustic information }
\label{ssec:acoustic}
Some methods used only acoustic information such as speech rate, intonation, pause duration etc., \cite{aud1, aud2}. While pauses influence in the prediction of Comma, intonation helps in disambiguation between punctuation marks like period and exclamation. Although this seemed to work, the most effective approach is to combine acoustic information with lexical information at word level using force-aligned duration \cite{klejch2017sequence}. In this work, we only considered lexical input and a pretrained lexical encoder for prediction of punctuation and truecasing. The use of pretrained acoustic encoder and fusion with lexical outputs are possible extensions in future work. 

\subsection{ Neural approaches}
\label{ssec:neural}
Neural approaches for punctuation and truecasing can be classified into two broad categories: sequence labeling based models and MT-based seq2seq models. These approaches have proven to be quite effective in capturing the contextual information and achieved huge success. While some approaches considered only punctuation prediction, some others jointly modeled punctuation and truecasing. 

One set of approaches treated punctuation as a machine translation problem and used phrase based statistical machine translation systems to output punctuated and true cased text \cite{seq1, seq2, seq3}. Inspired by recent end-to-end approaches, \cite{yi2019self} proposed the use of self-attention based transformer model to predict punctuation marks as output sequence for given word sequences. Most recently, \cite{nguyen2019fast} proposed joint modeling of punctuation and truecasing by generating words with punctuation marks as part of the decoding. Although seq2seq based approaches have shown a strong performance, they are intensive, demanding and are not suitable for production deployment at large scale. 

For sequence labeling problem, each word in the input is tagged with a punctuation. If there is no punctuation associated with a word, a blank label is used and is often referred as ``no punc''. \cite{cho2015comb} used a combination of neural networks and CRFs for joint prediction of punctuation and disfluencies. With growing popularity in deep recurrent neural networks, LSTMs and BLSTMs with attention mechanism were introduced for punctuation restoration \cite{ottokar2015lstm, ottokar2016blstm}. Later, \cite{pahuja2017joint} proposed joint training of punctuation and truecasing using BLSTM models. This work addressed joint learning as two correlated tasks, and predicted punctuation and truecasing as two independent outputs. Our proposed approach is similar to this work, but we rather condition truecasing prediction on punctuation output; this is discussed in detail in Section \ref{sec:proposed_framework}.

Punctuation and casing restoration for speech/ASR outputs in the medical domain has not been explored extensively. Recently, \cite{salloum2017deep} proposed a sequence labeling model using bi-directional RNNs with an attention mechanism and late fusion for punctuation restoration to clinical dictation. To our knowledge, there has not been any work on medical conversations, and we aim to bridge the gap here with latest advances in NLP with large-scale pretrained language models.

\begin{figure*}[t]
\begin{minipage}[b]{1.0\linewidth}
  \centering
  \centerline{\includegraphics[width=10.5cm]{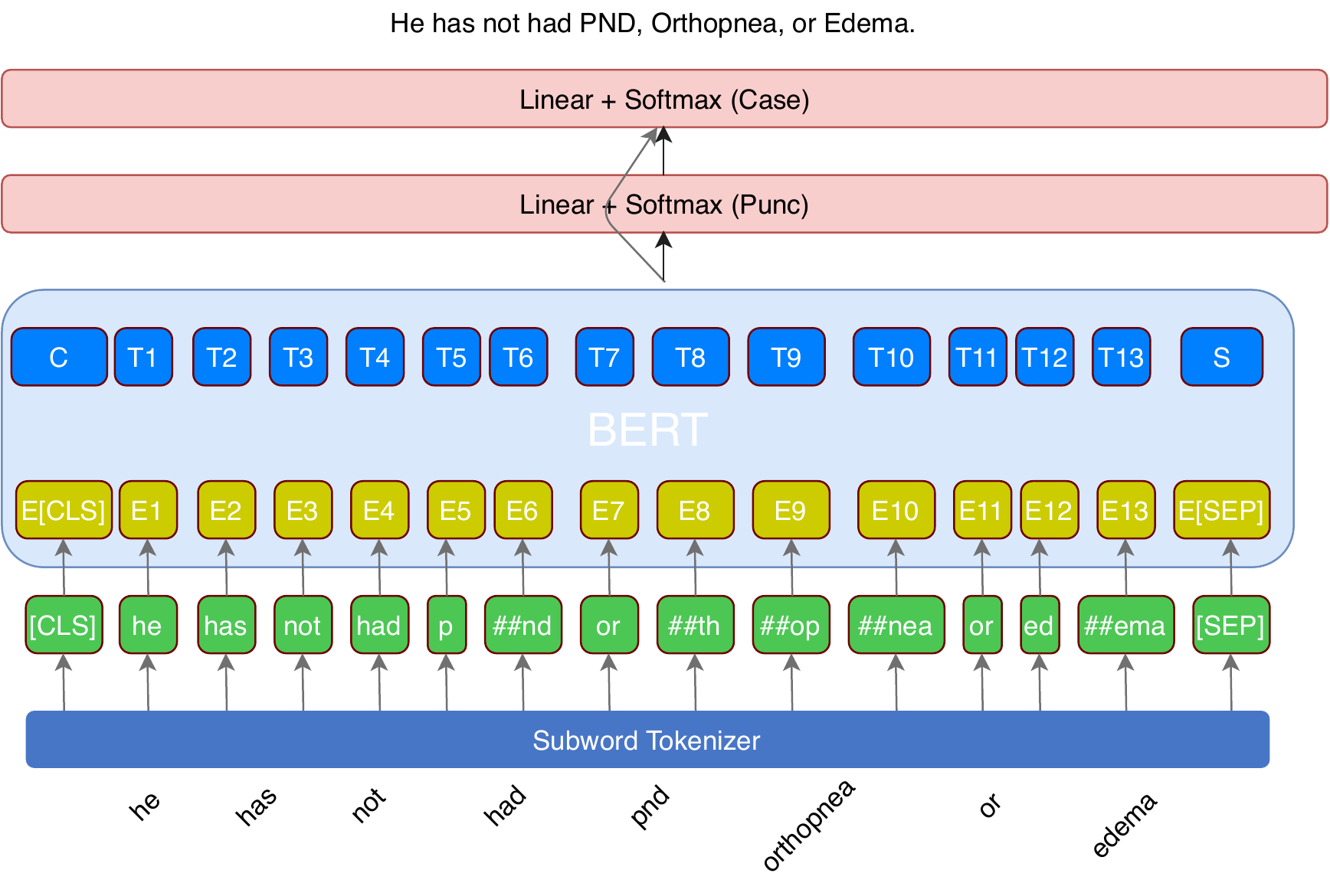}}
  \vspace{-2mm}
\caption{Pre-trained BERT encoder for prediction of punctuation and truecasing.}
\vspace{-4mm}
\label{fig:bert_punc}
\end{minipage}
\end{figure*}

\section{Modeling : Conditional Joint labeling of Punctuation + Casing}
\label{sec:proposed_framework}

We propose a postprocessing framework for conditional and joint learning of punctuation and truecasing prediction. Consider an input utterance $x_{1:T} = \{x_1, x_2,..., x_T\}$, of length $T$ and consisting of words $x_i$. The first step in our modeling process involves punctuation prediction as a sequence tagging task. Once the model predicts a probability distribution over punctuation, this along with the input utterance is fed in as input for predicting the case of a word $x_i$. We consider the punctuation to be independent of casing and a conditional dependence of the truecase of a word on punctuation given the learned input representations. Our plausible reasoning follows from this example sentence -- ``She took dance classes. She had no natural grace or sense of rhythm.''. The word after the period is capitalized, which implies that punctuation information can help in better prediction of casing. A pair of punctuation and truecasing is assigned per word: 
\begin{equation}
\begin{split}
    \mathbf{Pr(p_{1:T}, c_{1:T} | x_{1:T})}  = \hspace*{3cm}  \\
    \mathbf{Pr(p_{1:T}|x_{1:T})}  \mathbf{Pr(c_{1:T}|p_{1:T},x_{1:T})}
\end{split}
\end{equation}

where $c_i \in C$, a fixed set of casing labels \{Lower\_Case, Upper\_Case, All\_Caps, Mixed\_Case\}, and $p_i \in\ P$, a fixed set of punctuation labels \{Comma, Period, Question\_Mark, No\_Punct\}.

\subsection{Pretrained lexical encoder}

We propose to use a pretrained model like BERT, trained on a large text corpus, as a lexical encoder for learning an effective representation of the input utterance. Figure \ref{fig:bert_punc} illustrates our proposed model architecture.

\noindent\textbf{Subword embeddings} Given a sequence of input vectors ($x_1, x_2,..., x_T$), where $x_i$ represents a word $w_i$, we extract the subword embeddings ($s_1, s_2,..., s_n$) using a wordpiece tokenizer \cite{wordpiece}. Using subwords is especially effective in medical domain, as it contains more compound words with common subwords. For example consider the six words \{hypotension, hypertension, hypoactive, hyperactive, active, tension \} with four common subwords \{hyper, hypo, active, tension\}. In Section \ref{ssec:word_vs_suwbword}, we provide a comparative analysis of word and subword models across different architectures on medical data.

\noindent\textbf{BERT encoder} We provide subword embeddings ($s_1, s_2,...,s_n$) as input to the BERT encoder, which outputs a sequence of hidden states: H = ($h1, . . . , h_n$ ) at its final layer. The pretrained BERT base encoder consists of 12 transformer encoder self-attention layers. For this task, we truncate the BERT encoder and fine-tune only the first six layers to reduce the model complexity. Although a deep encoder might enable us to learn a long memory context dependent representation of the input utterance, the performance gain is very minimal compared to the increased latency\footnote{We experimentally found that 12-layer BERT base model gives $\sim$1\% improvement over 6-layer BERT base model whereas the inference and training times were double for the former.}.

For punctuation, we input the last layer representations of truncated BERT encoder $h_1, h_2,..., h_n$ to a linear layer with softmax activation to classify over the punctuation labels generating ($p_1, p_2,..., p_n$) as outputs. For casing, we concatenate the softmax probabilities of punctuation output with BERT encoder's outputs and feed to a linear layer with softmax activation generating case labels ($c_1, c_2,..., c_n$) for the sequence. The softmax output for punctuation ($\hat{p_i}$) and truecasing ($\hat{c_i}$) is as follows:

\begin{equation}
    \hat{p_i} = softmax(W^kh_i + b^k) 
\end{equation}
\begin{equation}
    \hat{c_i} = softmax(W^l(\hat{p_i} \oplus h_i) + b^l) 
\end{equation}   

where $W^k$, $b^k$ denote weights and bias of punctuation linear output layer and $W^l$, $b^l$ denote weights and bias of truecasing linear output layer. 

\noindent\textbf{Joint learning objective: } We model our learning objective to maximize the joint probability $\mathbf{Pr(p_{1:T}, c_{1:T} | x_{1:T})}$. The model is finetuned end-to-end to minimize the cross-entropy loss between the assigned distribution and the training data. The parameters of BERT encoder are shared across punctuation and casing prediction tasks and are jointly trained. We compute the losses ($L^p, L^c$) for each task using cross entropy loss function.  The final loss L to be optimized is a weighted average of the task-specific loses:
\begin{equation}
    L = \alpha L^p + L^c
\end{equation} 

where $\alpha$ is a fixed weight optimized for best predictions across both the tasks. In our experiments, we explored  $\alpha$ values in the range of (0.2-2) and found 0.6 to be the optimal value.

\subsection{Finetuning using Masked Language Model with Medical domain data} 
\label{ssec:finetuning_mlm}

BERT and its successors have shown great performance on downstream NLP tasks. But just like any other model, these Language Models are biased by their training data. In particular, they are typically trained on data that is easily available in large quantities on the internet e.g. Wikipedia, Common-Crawl etc. Our domain, Medical ASR Text, is not ``common'' and is very under-represented in the training data for these Language Models. One way to correct this situation is to perform a few steps of unsupervised Masked Language Model finetuning on the BERT models before performing cross-entropy training using the labeled task data \cite{han2019unsupervised}.

\noindent\textbf{Domain adaptation} We finetune the pretrained BERT model for MLM (Masked LM) objective on medical domain data. 15\% of input tokens are masked randomly before feeding into the BERT model as proposed by \cite{bert}. The main goal is to adapt and learn better representations of speech data. The domain adapted model can be
further finetuned with an additional layer to a downstream task like punctuation and casing prediction.

\noindent\textbf{Domain+Task adaptation} Building on the previous technique, we attempt to finetune the pretrained model for task adaptation in combination with domain adaptation. In this technique, instead of randomly masking 15\% of the input tokens, we do selective masking i.e. 50\% of the masked tokens would be random and the other 50\% would be punctuation marks ([``.'', ``,'', ``?''] in our case). Therefore, the finetuned model would not only adapt to speech domain, but would also effectively learn the placement of punctuation marks in a text based on the context.

\begin{table*}[t]
\begin{tabular}{lcccccccc}
\hline
                    & \multicolumn{1}{l}{} & \multicolumn{3}{c}{Punctuation} & \multicolumn{4}{c}{Truecasing} \\
Model               & Token   & No Punc & Full stop & Comma & LC   & UC   & CA   & MC   \\ \hline
CNN-Highway         & word    & 0.97    & 0.81      & 0.71  & 0.98 & 0.84 & 0.95 & 0.99 \\
                    & subword &   0.98      &  0.83  & 0.70      & 0.99     &  0.87    & 0.95     & 0.99     \\
3-LSTM              & word    & 0.97    & 0.82      & 0.73  & 0.98 & 0.84 & 0.96 & 0.98 \\
                    & subword & 0.98    & 0.84      & 0.75  & 0.99 & 0.87 & 0.97 & 0.99 \\
3-BLSTM             & word    & 0.98    & 0.86      & 0.75  & 0.99 & 0.88 & 0.97 & 0.98 \\
                    & subword & 0.99    & 0.87      & 0.76  & 0.99 & 0.90 & 0.97 & 1.0  \\
Transformer encoder & word    & 0.97 &	0.84 &	0.7 &	0.98 &	0.86 &	0.97 &	0.98     \\
                    & subword & 0.98 &	0.85 &	0.72 &	0.99 &	0.87 &	0.97 &	0.99 \\
\hline
\end{tabular}
\caption{Dictation corpus: Comparison of F1 scores for punctuation and truecasing across different model architectures using word and subword tokens (LC: lower case; UC: Upper case; CA: CAPS All; MC: Mixed Case).}
\vspace{-2mm}
\label{tab:dictationwordvsubword}
\end{table*}

\begin{table*}[t]
\begin{tabular}{lccccccccc}
\hline
                    & \multicolumn{1}{l}{} & \multicolumn{4}{c}{Punctuation} & \multicolumn{4}{c}{Truecasing} \\
Model               & Token   & No Punc & Full stop & Comma & QM & LC   & UC   & CA   & MC   \\ \hline
CNN-Highway         & word    & 0.96    & 0.72 & 0.64 & 0.60 & 0.96 & 0.78 & 0.99 & 0.91 \\
                    & subword & 0.97 & 0.74 & 0.65 & 0.61 & 0.97 & 0.80 & 0.98 & 0.99     \\
3-LSTM              & word  & 0.96 &0.74 &0.64 &0.65 &0.96 &0.79 &0.99 &0.95 \\
                    & subword & 0.97 & 0.75 & 0.65 & 0.66 & 0.97 &0.79 &0.97 & 1.0 \\
3-BLSTM             & word    & 0.97 & 0.77 &0.68 &0.68 &0.97 &0.82 &0.99 &0.95 \\
                    & subword & 0.98 & 0.79 &0.68 &0.69 &0.97 &0.83 &0.99 & 1.0 \\
Transformer encoder & word    &  0.97 &	0.77 &	0.68 &	0.68 &	0.97 &	0.83 &	0.99 &	0.92 \\
             & subword & 0.98 &	0.79 &	0.69 &	0.69 & 0.98 &	0.83 &	0.99 &	1.0 \\
\hline
\end{tabular}
\caption{Conversational corpus: Comparison of F1 scores for punctuation and truecasing across different model architectures using word and subword tokens (QM: Question Mark; LC: lower case; UC: Upper case; CA: CAPS All; MC: Mixed Case).}
\vspace{-4mm}
\label{tab:convwordvsubword}
\end{table*}

\subsection{Robustness to ASR errors}
\label{ssec:robustness_asr}

 Models trained on ground truth text inputs may not perform well when tested with ASR output, especially when the system introduces grammatical errors. To make models more robust against ASR errors, we perform data augmentation with ASR outputs for training. For punctuation restoration, we use edit distance measure to align ASR hypothesis with ground truth punctuated text. Before computing alignment, we strip all punctuation from ground truth and lowercase the text. This helps us find the best alignment between ASR hypothesis and ground truth text. Once the alignment is found, we restore the punctuation from each word in ground truth text to hypothesis. If there are words that are punctuated in ground truth but got deleted in ASR hypothesis, we restore the punctuation to previous word. For truecasing, we try to match the reference word with hypothesis word from aligned sequences with a window size of 5, two words to the left and two words to the right of current word and restore truecasing only in the cases where reference word is found. We performed experiments with data augmentation using 1-best hypothesis and n-best lists as additional training data and the results are reported in Section \ref{ssec:robustness}.

\begin{table*}[t]
\begin{tabular}{lcccccccc}
\hline
                    & \multicolumn{1}{l}{} & \multicolumn{3}{c}{Punctuation} & \multicolumn{4}{c}{Truecasing} \\
Model               & Dataset   & No Punc & Full stop & Comma & LC   & UC   & CA   & MC   \\ \hline
3-BLSTM     & Wiki    & 0.95    & 0.17    & 0.27  & 0.95 & 0.31 & 0.55 & 0.19 \\
BERT    & Wiki &   0.96      &  0.2  & 0.39      & 0.95     &  0.36    & 0.65     & 0.2     \\ \hline
3-BLSTM              & Medical    & 0.99    & 0.87      & 0.76  & 0.99 & 0.9 & 0.97 & 1.0 \\
BERT                    & Medical & 0.99    & 0.9      & 0.81  & 0.99 & 0.93 & 0.99 & 1.0 \\
FT-BERT & Medical & 0.99 &	0.92 &	0.82 &	0.99 &	0.93 &	0.99 &	1.0 \\
PM-BERT & Medical & 0.99 &	0.93 &	0.82 &	0.99 &	0.94 &	0.99 &	1.0  \\ \hline
Bio-BERT             & Medical    & 0.99    & 0.92      & 0.82  & 0.99 & 0.93 & 0.99 & 1.0 \\
RoBERTa                    & Medical & 0.99    & 0.92      & 0.81  & 0.99 & 0.94 & 0.99 & 1.0  \\ 
\hline
\end{tabular}
\caption{Comparison of F1 scores for punctuation and truecasing using BERT and BLSTM when trained on Wiki data and Medical dictation data (FT-BERT: Finetuned BERT for domain adapation, PM-BERT: Finetuned BERT by punctuation masking for domain and task adapation).}
\vspace{-2mm}
\label{tab:dictationbert}
\end{table*}

\begin{table*}[t]
\begin{tabular}{lccccccccc}
\hline
                    & \multicolumn{1}{l}{} & \multicolumn{4}{c}{Punctuation} & \multicolumn{4}{c}{Truecasing} \\
Model               & Dataset   & No Punc & Full stop & Comma & QM & LC   & UC   & CA   & MC   \\ \hline
3-BLSTM     & Wiki    & 0.89    & 0.001    & 0.25  &0.002 & 0.93 & 0.13 & 0.9 & 0.95 \\
BERT    & Wiki &   0.93	& 0.004	& 0.4 &	0.007 &	0.93 &	0.4 &	0.95 &	0.95  \\ \hline
3-BLSTM              & Medical    & 0.98 &	0.79 &	0.68 &	0.69 &	0.97 &	0.83 &	0.99 &	1.0
 \\
BERT                    & Medical & 0.98 &	0.8 &	0.71 &	0.72 &	0.98 &	0.85 &	0.99 &	1.0
 \\
FT-BERT & Medical & 0.98 &	0.81 &	0.72 &	0.73 &	0.98 &	0.85 &	0.99 &	1.0  \\
PM-BERT & Medical & 0.98 &	0.82 &	0.72 &	0.74 &	0.98 &	0.86 &	0.99 &	1.0  \\ \hline
Bio-BERT             & Medical    & 0.98 &	0.81 &	0.71 &	0.72 &	0.98 &	0.85 &	0.99 &	1.0 \\
RoBERTa                    & Medical &  0.98 &	0.82 &	0.73 &	0.74 &	0.98 &	0.86 &	0.99 &	1.0  \\
\hline
\end{tabular}
\caption{Comparison of F1 scores for punctuation and truecasing using BERT and BLSTM when trained on Wiki data and Medical conversation data (FT-BERT: Finetuned BERT for domain adapation, PM-BERT: Finetuned BERT by punctuation masking for domain and task adapation).}
\vspace{-4mm}
\label{tab:conversationbert}
\end{table*}

\section{Experiments and results}
\label{sec:experiments}

\subsection {Data}
We evaluate our proposed framework and models on a subset of two internal medical datasets: dictation and conversational. The dictation corpus contains 3.7M words and the conversational corpus contains 51M words. The medical data comes with special tags masking personal identifiable and patient health information. We also use a general domain Wikipedia dataset for comparative analysis with Medical domain data. This data is a subset of the publicly available release of Wiki dataset \cite{sproatWiki}. The corpus contains 35M words and relatively shorter sentences ranging from 8 to 200 words in length. 90\% of the data from each corpus is used for training, 5\% for fine-tuning and remaining 5\% is held-out for testing. 

For robustness experiments presented in Section \ref{ssec:robustness}, we used data from the dictation corpus consisting of 2265 text files and corresponding audio files with an average duration of $\sim$15 minutes. The total length of the corpus is 550 hours. For augmentation with ground-truth transcription, we transcribed audio files using a speech recognition system. Restoration of punctuation and truecasing to transcribed text can be erroneous as the word error rate(\textit{WER}) goes up. We therefore discarded the transcribed text of those audio files whose WER is more than 25\%. We sorted the remaining transcriptions based on WER to make further splits: hypothesis from top 50 files with best WER is set as test data, and the next 50 files were chosen as development and rest of the transcribed text was used for training. The partition was done this way to minimize the number of errors that may occur during restoration. 

\noindent \textbf{Preprocessing long-speech transcriptions} 
Conversational style speech has long-speech transcripts, in which the context is spread across multiple segments. we use an overlapped chunking and merging component to pre and post process the data. We use a sliding window approach \cite{chunk} to split long ASR outputs into chunks of 200 words each with an overlapping window of 50 words each to the left and right. The overlap helps in preserving the context for all the words after splitting and ensures accurate prediction of punctuation and case corresponding to each word. 

\subsection{Large Vocabulary: Word vs Subword models}
\label{ssec:word_vs_suwbword}
For a fair comparison with BERT, we evaluate various recurrent and non-recurrent architectures with both word and subword embeddings. The two recurrent models include a 3 layer uni-directional LSTM (\textit{3-LSTM}) and a 3 layer Bi-directional LSTM (\textit{3-BLSTM}). One of the non recurrent encoders, implements a CNN-Highway architecture based on the work proposed by \cite{cnnhighway}, whereas the other one implements a transformer encoder based model \cite{attention}. We train all four models on medical data from dictation and conversation corpus with weights initialized randomly. The vocabulary for word models is derived by considering all the unique words from training corpus, with additional tokens for unknown and padding. This yielded a vocabulary size of 30k for dictation and 64k for conversational corpus. Subwords are extracted using a wordpiece model \cite{wordpiece} and its inventory is less than half that of word model for conversation. Tables \ref{tab:dictationwordvsubword} and \ref{tab:convwordvsubword} summarize our results on dictation and conversation datasets respectively. We observe that subword models consistently performed same or better than word models. On punctuation task, for Full stop and Comma, we notice an absolute $\sim$1-2\% improvement respectively on dictation set. Similarly, on the conversation dataset, we notice an absolute $\sim$1-2\% improvement on Full stop, Comma and Question Mark. For the casing task, we notice that word and subword models performed equally well except in dictation dataset where we see an absolute $\sim$3\% improvement for Upper\_Case. We hypothesize that medical vocabulary contains a large set of compound words, which a subword based model works effectively over word model. Upon examining few utterances, we noticed that subword models can learn effective representations of these compound medical words by tokenizing them into subwords. On the other hand, word models often run into rare word or OOV issues.

\begin{table*}[t]
\begin{tabular}{lccccc||cccc}
\hline
                    & \multicolumn{1}{l}{} & \multicolumn{4}{c}{Punctuation} & \multicolumn{4}{c}{Truecasing}  
                    \\
Model       & n-best         & No Punc & Full stop & Comma & QM & LC   & UC   & CA   & MC  \\ \hline
BERT-GT     & -       & 0.97    & 0.58 & 0.45 & 0.0 & 0.98 & 0.60 & 0.78 & 0.90  \\
BERT-ASR  & 1-best  & 0.97    & 0.66 & 0.56 & 0.54 & 0.99 & 0.72 & 0.86 & 1.0 \\
  & 3-best  & 0.98 &	0.67 &	0.57 &	0.42 &	0.98 &	0.69 &	0.79 &	0.84 \\
  & 5-best  & 0.97 &	0.61 &	0.5	 & 0.35	& 0.98 &	0.65 &	0.79 &	0.83 \\
\hline
\end{tabular}
\caption{Comparison of F1 scores for punctuation and truecasing with ground truth and ASR augmented data.}
\vspace{-4mm}
\label{tab:robustness}
\end{table*}

\subsection{Pretrained language models}
\label{ssec:data_scarcity}

\noindent\textbf {Significance of in-domain data} For analyzing the importance of in-domain data, we train a baseline BLSTM model and a pretrained BERT model on Wiki and Medical data from both dictation and conversational corpus and tested the models on Medical held-out data. The first four rows of Tables  \ref{tab:dictationbert} and \ref{tab:conversationbert} summarize the results. The models trained on Wiki data performed very poorly when compared to models trained on Medical data from either dictation or conversation corpus. Although dictation corpus (3.7M words) is relatively smaller than Wiki corpus (35M words), the difference in accuracy is significantly higher across both models. Imbalanced classes like Full stop, Comma, Question\_Mark were most affected. Another interesting observation is that the models trained on Medical data performed better on Full stop compared to Comma; whereas general domain models performed better on Comma compared to Full stop. The degradation in general models might be due to Wiki sentences being short and ending with a Full stop unlike lengthy medical transcripts. Also, the F1 scores are lower on conversation data across both the tasks, indicating the complexity involved in modeling conversational data due to their highly unstructured format. Overall, the pretrained BERT model consistently outperformed baseline BLSTM model on both dictation and conversation data. This motivated us to focus on adapting the pretrained models for this task. 
\\ \\
\noindent\textbf{Finetuning Masked LM} We have run two levels of fine-tuning as explained in Section \ref{ssec:finetuning_mlm}. First, we finetuned BERT with Medical domain data using random masking (\textit{FT-BERT}) and for task adaptation, we performed fine-tuning with punctuation based masking (\textit{PM-BERT}). For both experiments, we used the same data as we have used for finetuning the downstream task. From the results presented in Table \ref{tab:dictationbert} and \ref{tab:conversationbert}, we infer that finetuning boosts the performance of punctuation and truecasing (an absolute improvement of $\sim$1-2\%). From both the datasets, it is clear that task specific masking helps better than simple random masking. For dictation dataset, Full stop improved by an absolute 3\% by performing punctuation specific masking, suggesting that finetuning MLM can give higher benefits when the amount of data is low.

\begin{figure}[t]
\begin{minipage}[b]{1.0\linewidth}
  \centering
  \centerline{\includegraphics[width=7.5cm]{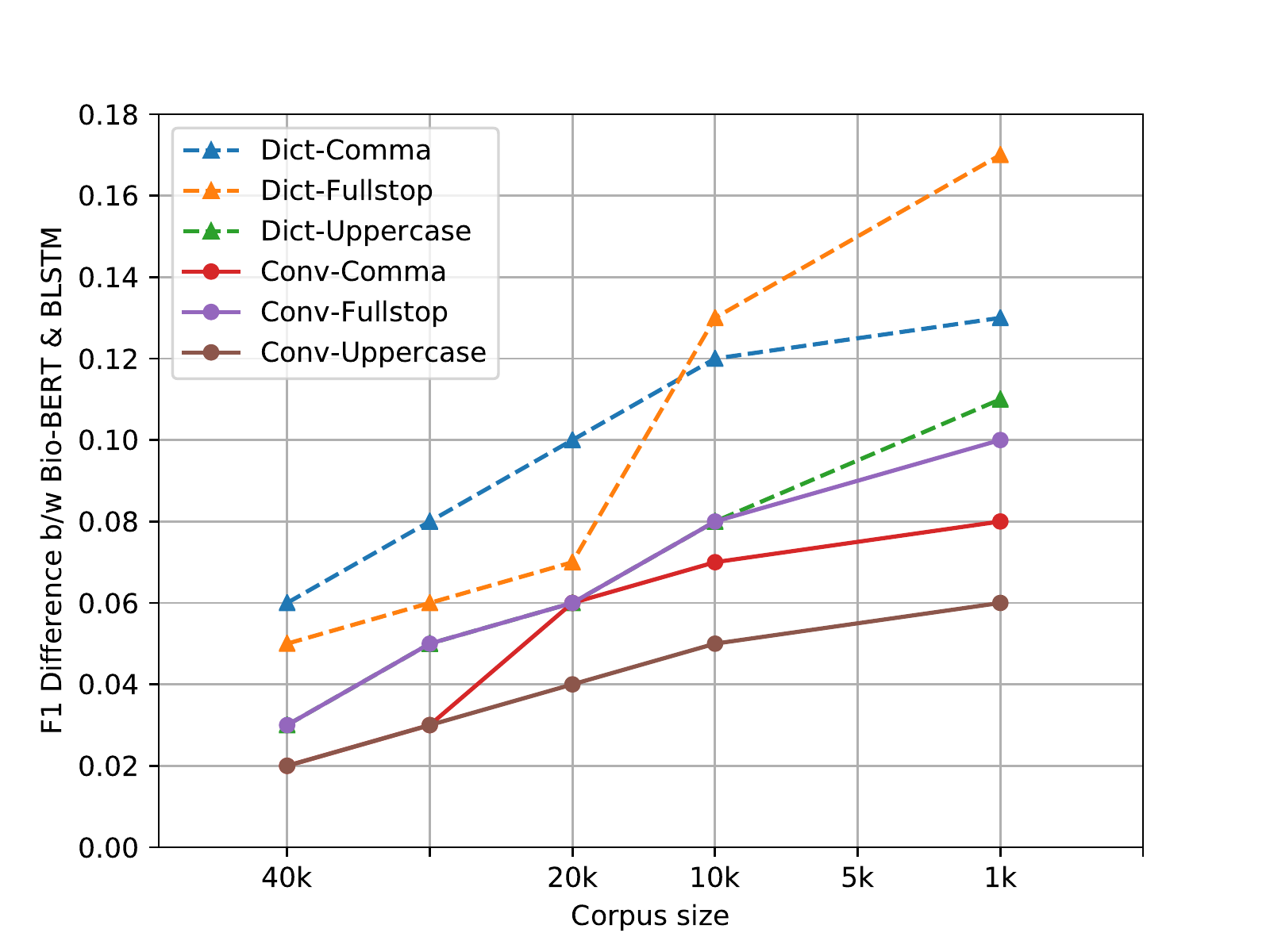}}
  \vspace{-2mm}
\caption{Difference in F1 scores between Bio-BERT and BLSTM for varying data sizes.}
\vspace{-4mm}
\label{fig:bert}
\end{minipage}
\end{figure}

\noindent\textbf {Variants of BERT} We compare three pretrained models namely, BERT and its successor RoBERTa \cite{roberta} and Bio-BERT \cite{lee2020biobert} which was trained on large scale Biomedical corpora. The results are summarized in last two rows of Table \ref{tab:dictationbert} and \ref{tab:conversationbert}. First, we observe that both Bio-BERT and RoBERTa outperformed the initial BERT model and has shown an absolute $\sim$3-5\% improvement over the baseline 3-BSLTM. To further validate this, we extended our experiments to understand how the performance of our best model(Bio-BERT) varies across different training dataset sizes compared to the baseline. From Figure \ref{fig:bert}, we observe that the difference increases significantly as we move towards smaller datasets. For the smallest data set size of ~500k words (1k transcripts), there is an absolute improvement of 6-17\% over the baseline in accuracy in terms of F1. This shows that pretraining on a large dataset helps to overcome data scarcity issue effectively. 

\subsection{Robustness}
\label{ssec:robustness}

For testing robustness, we performed experiments with augmentation of ASR data from n-best lists (\textit{BERT-ASR}). We considered top-1, top-3 and top-5 hypotheses for n-best lists augmentation with ground truth text and the results are presented in Table \ref{tab:robustness}.  Additionally, the best BERT model trained using only ground truth text inputs (\textit{BERT-GT}) from Table \ref{tab:dictationbert} is also evaluated on ASR outputs. To compute F1 scores on held-out test set, we first aligned the ASR hypothesis with ground truth data and restored the punctuation and truecasing as described in Section \ref{ssec:robustness_asr}. From the results presented in Table \ref{tab:robustness}, we infer that adding ASR hypothesis to the training data helped improve the performance of both punctuation and truecasing. In punctuation, both Full stop and Comma have seen an absolute 10\% improvement in F1 score. Although the number of question marks is less in test data, the augmented systems performed really well compared to the system trained purely on ground truth text. However, we found that using n-best lists with $n>1$ did not help much compared to the 1-best list. This may be due to sub-optimal restoration of punctuation and truecasing as the WER with n-best lists is likely to go up as $n$ increases.

\section{Conclusion}
\label{sec:conclusion}

In this paper, we have presented a framework for conditional joint modeling of punctuation and truecasing in medical transcriptions using pretrained language models such as BERT. We also demonstrated the benefit from MLM objective finetuning of the pretrained model with task specific masking. We further improved the robustness of punctuation and truecasing on ASR outputs by data augmentation during training. Experiments performed on both dictation and conversation corpora show the effectiveness of the proposed approach. Future work includes the use of either pretrained acoustic features or pretrained acoustic encoder to perform fusion with pretrained linguistic encoder to further boost the performance of punctuation. 


\bibliography{acl2020}

\begin{thebibliography}{37}
\expandafter\ifx\csname natexlab\endcsname\relax\def\natexlab#1{#1}\fi

\bibitem[{Beeferman et~al.(1998)Beeferman, Berger, and Lafferty}]{lm3}
Doug Beeferman, Adam Berger, and John Lafferty. 1998.
\newblock Cyberpunc: A lightweight punctuation annotation system for speech.
\newblock In \emph{Proceedings of the 1998 IEEE International Conference on
  Acoustics, Speech and Signal Processing, ICASSP'98 (Cat. No. 98CH36181)},
  volume~2, pages 689--692. IEEE.

\bibitem[{Bodapati et~al.(2019)Bodapati, Gella, Bhattacharjee, and
  Al-Onaizan}]{bodapati-etal-2019-neural}
Sravan Bodapati, Spandana Gella, Kasturi Bhattacharjee, and Yaser Al-Onaizan.
  2019.
\newblock \href {https://doi.org/10.18653/v1/W19-3515} {Neural word
  decomposition models for abusive language detection}.
\newblock In \emph{Proceedings of the Third Workshop on Abusive Language
  Online}, pages 135--145, Florence, Italy. Association for Computational
  Linguistics.

\bibitem[{Chiu et~al.(2017)Chiu, Tripathi, Chou, Co, Jaitly, Jaunzeikare,
  Kannan, Nguyen, Sak, Sankar et~al.}]{ConvASR}
Chung-Cheng Chiu, Anshuman Tripathi, Katherine Chou, Chris Co, Navdeep Jaitly,
  Diana Jaunzeikare, Anjuli Kannan, Patrick Nguyen, Hasim Sak, Ananth Sankar,
  et~al. 2017.
\newblock Speech recognition for medical conversations.
\newblock \emph{arXiv preprint arXiv:1711.07274}.

\bibitem[{Cho et~al.(2015)Cho, Kilgour, Niehues, and Waibel}]{cho2015comb}
Eunah Cho, Kevin Kilgour, Jan Niehues, and Alex Waibel. 2015.
\newblock Combination of nn and crf models for joint detection of punctuation
  and disfluencies.
\newblock In \emph{Sixteenth annual conference of the international speech
  communication association}.

\bibitem[{Cho et~al.(2012)Cho, Niehues, and Waibel}]{seq2}
Eunah Cho, Jan Niehues, and Alex Waibel. 2012.
\newblock Segmentation and punctuation prediction in speech language
  translation using a monolingual translation system.
\newblock In \emph{International Workshop on Spoken Language Translation
  (IWSLT) 2012}.

\bibitem[{Christensen et~al.(2001{\natexlab{a}})Christensen, Gotoh, and
  Renals}]{fst2}
Heidi Christensen, Yoshihiko Gotoh, and Steve Renals. 2001{\natexlab{a}}.
\newblock Punctuation annotation using statistical prosody models.
\newblock In \emph{ISCA tutorial and research workshop (ITRW) on prosody in
  speech recognition and understanding}.

\bibitem[{Christensen et~al.(2001{\natexlab{b}})Christensen, Gotoh, and
  Renals}]{aud1}
Heidi Christensen, Yoshihiko Gotoh, and Steve Renals. 2001{\natexlab{b}}.
\newblock Punctuation annotation using statistical prosody models.
\newblock In \emph{ISCA tutorial and research workshop (ITRW) on prosody in
  speech recognition and understanding}.

\bibitem[{Devlin et~al.(2018)Devlin, Chang, Lee, and Toutanova}]{bert}
Jacob Devlin, Ming-Wei Chang, Kenton Lee, and Kristina Toutanova. 2018.
\newblock Bert: Pre-training of deep bidirectional transformers for language
  understanding.
\newblock \emph{arXiv preprint arXiv:1810.04805}.

\bibitem[{Driesen et~al.(2014)Driesen, Birch, Grimsey, Safarfashandi, Gauthier,
  Simpson, and Renals}]{seq3}
Joris Driesen, Alexandra Birch, Simon Grimsey, Saeid Safarfashandi, Juliet
  Gauthier, Matt Simpson, and Steve Renals. 2014.
\newblock Automated production of true-cased punctuated subtitles for weather
  and news broadcasts.
\newblock In \emph{Fifteenth Annual Conference of the International Speech
  Communication Association}.

\bibitem[{Edwards et~al.(2017)Edwards, Salloum, Finley, Fone, Cardiff, Miller,
  and Suendermann-Oeft}]{DictASR}
Erik Edwards, Wael Salloum, Greg~P Finley, James Fone, Greg Cardiff, Mark
  Miller, and David Suendermann-Oeft. 2017.
\newblock Medical speech recognition: reaching parity with humans.
\newblock In \emph{International Conference on Speech and Computer}, pages
  512--524. Springer.

\bibitem[{Gotoh and Renals(2000)}]{fst1}
Yoshihiko Gotoh and Steve Renals. 2000.
\newblock Sentence boundary detection in broadcast speech transcripts.

\bibitem[{Gravano et~al.(2009)Gravano, Jansche, and Bacchiani}]{lm5}
Agustin Gravano, Martin Jansche, and Michiel Bacchiani. 2009.
\newblock Restoring punctuation and capitalization in transcribed speech.
\newblock In \emph{2009 IEEE International Conference on Acoustics, Speech and
  Signal Processing}, pages 4741--4744. IEEE.

\bibitem[{Han and Eisenstein(2019)}]{han2019unsupervised}
Xiaochuang Han and Jacob Eisenstein. 2019.
\newblock Unsupervised domain adaptation of contextualized embeddings for
  sequence labeling.
\newblock In \emph{Proceedings of the 2019 Conference on Empirical Methods in
  Natural Language Processing and the 9th International Joint Conference on
  Natural Language Processing (EMNLP-IJCNLP)}, pages 4229--4239.

\bibitem[{Huang and Zweig(2002)}]{mem1}
Jing Huang and Geoffrey Zweig. 2002.
\newblock Maximum entropy model for punctuation annotation from speech.
\newblock In \emph{Seventh International Conference on Spoken Language
  Processing}.

\bibitem[{Kim et~al.(2016)Kim, Jernite, Sontag, and Rush}]{cnnhighway}
Yoon Kim, Yacine Jernite, David Sontag, and Alexander~M Rush. 2016.
\newblock Character-aware neural language models.
\newblock In \emph{AAAI}, pages 2741--2749.

\bibitem[{Klejch et~al.(2017)Klejch, Bell, and Renals}]{klejch2017sequence}
Ond{\v{r}}ej Klejch, Peter Bell, and Steve Renals. 2017.
\newblock Sequence-to-sequence models for punctuated transcription combining
  lexical and acoustic features.
\newblock In \emph{2017 IEEE International Conference on Acoustics, Speech and
  Signal Processing (ICASSP)}, pages 5700--5704. IEEE.

\bibitem[{Lee et~al.(2020)Lee, Yoon, Kim, Kim, Kim, So, and
  Kang}]{lee2020biobert}
Jinhyuk Lee, Wonjin Yoon, Sungdong Kim, Donghyeon Kim, Sunkyu Kim, Chan~Ho So,
  and Jaewoo Kang. 2020.
\newblock Biobert: a pre-trained biomedical language representation model for
  biomedical text mining.
\newblock \emph{Bioinformatics}, 36(4):1234--1240.

\bibitem[{Levy et~al.(2012)Levy, Silber-Varod, and Moyal}]{aud2}
Tal Levy, Vered Silber-Varod, and Ami Moyal. 2012.
\newblock The effect of pitch, intensity and pause duration in punctuation
  detection.
\newblock In \emph{2012 IEEE 27th Convention of Electrical and Electronics
  Engineers in Israel}, pages 1--4. IEEE.

\bibitem[{Liu et~al.(2019)Liu, Ott, Goyal, Du, Joshi, Chen, Levy, Lewis,
  Zettlemoyer, and Stoyanov}]{roberta}
Yinhan Liu, Myle Ott, Naman Goyal, Jingfei Du, Mandar Joshi, Danqi Chen, Omer
  Levy, Mike Lewis, Luke Zettlemoyer, and Veselin Stoyanov. 2019.
\newblock Roberta: A robustly optimized bert pretraining approach.
\newblock \emph{arXiv preprint arXiv:1907.11692}.

\bibitem[{Lu and Ng(2010)}]{CRF2}
Wei Lu and Hwee~Tou Ng. 2010.
\newblock Better punctuation prediction with dynamic conditional random fields.
\newblock In \emph{Proceedings of the 2010 conference on empirical methods in
  natural language processing}, pages 177--186.

\bibitem[{Makhoul et~al.(2005)Makhoul, Baron, Bulyko, Nguyen, Ramshaw,
  Stallard, Schwartz, and Xiang}]{nlp22}
John Makhoul, Alex Baron, Ivan Bulyko, Long Nguyen, Lance Ramshaw, David
  Stallard, Richard Schwartz, and Bing Xiang. 2005.
\newblock The effects of speech recognition and punctuation on information
  extraction performance.
\newblock In \emph{Ninth European Conference on Speech Communication and
  Technology}.

\bibitem[{Nguyen et~al.(2019{\natexlab{a}})Nguyen, Nguyen, Nguyen, Phuong,
  Nguyen, Do, and Mai}]{chunk}
Binh Nguyen, Vu~Bao~Hung Nguyen, Hien Nguyen, Pham~Ngoc Phuong, The-Loc Nguyen,
  Quoc~Truong Do, and Luong~Chi Mai. 2019{\natexlab{a}}.
\newblock Fast and accurate capitalization and punctuation for automatic speech
  recognition using transformer and chunk merging.
\newblock \emph{arXiv preprint arXiv:1908.02404}.

\bibitem[{Nguyen et~al.(2019{\natexlab{b}})Nguyen, Nguyen, Nguyen, Phuong,
  Nguyen, Do, and Mai}]{nguyen2019fast}
Binh Nguyen, Vu~Bao~Hung Nguyen, Hien Nguyen, Pham~Ngoc Phuong, The-Loc Nguyen,
  Quoc~Truong Do, and Luong~Chi Mai. 2019{\natexlab{b}}.
\newblock Fast and accurate capitalization and punctuation for automatic speech
  recognition using transformer and chunk merging.
\newblock \emph{arXiv preprint arXiv:1908.02404}.

\bibitem[{Pahuja et~al.(2017)Pahuja, Laha, Mirkin, Raykar, Kotlerman, and
  Lev}]{pahuja2017joint}
Vardaan Pahuja, Anirban Laha, Shachar Mirkin, Vikas Raykar, Lili Kotlerman, and
  Guy Lev. 2017.
\newblock Joint learning of correlated sequence labelling tasks using
  bidirectional recurrent neural networks.
\newblock \emph{arXiv preprint arXiv:1703.04650}.

\bibitem[{Peitz et~al.(2011{\natexlab{a}})Peitz, Freitag, Mauser, and
  Ney}]{nlp11}
Stephan Peitz, Markus Freitag, Arne Mauser, and Hermann Ney.
  2011{\natexlab{a}}.
\newblock Modeling punctuation prediction as machine translation.
\newblock In \emph{International Workshop on Spoken Language Translation
  (IWSLT) 2011}.

\bibitem[{Peitz et~al.(2011{\natexlab{b}})Peitz, Freitag, Mauser, and
  Ney}]{seq1}
Stephan Peitz, Markus Freitag, Arne Mauser, and Hermann Ney.
  2011{\natexlab{b}}.
\newblock Modeling punctuation prediction as machine translation.
\newblock In \emph{International Workshop on Spoken Language Translation
  (IWSLT) 2011}.

\bibitem[{Salloum et~al.(2017)Salloum, Finley, Edwards, Miller, and
  Suendermann-Oeft}]{salloum2017deep}
Wael Salloum, Gregory Finley, Erik Edwards, Mark Miller, and David
  Suendermann-Oeft. 2017.
\newblock Deep learning for punctuation restoration in medical reports.
\newblock In \emph{BioNLP 2017}, pages 159--164.

\bibitem[{Schuster and Nakajima(2012)}]{wordpiece}
Mike Schuster and Kaisuke Nakajima. 2012.
\newblock Japanese and korean voice search.
\newblock In \emph{2012 IEEE International Conference on Acoustics, Speech and
  Signal Processing (ICASSP)}, pages 5149--5152. IEEE.

\bibitem[{Sennrich et~al.(2015)Sennrich, Haddow, and Birch}]{subwordnmt}
Rico Sennrich, Barry Haddow, and Alexandra Birch. 2015.
\newblock Neural machine translation of rare words with subword units.
\newblock \emph{arXiv preprint arXiv:1508.07909}.

\bibitem[{Sproat and Jaitly(2016)}]{sproatWiki}
Richard Sproat and Navdeep Jaitly. 2016.
\newblock Rnn approaches to text normalization: A challenge.
\newblock \emph{arXiv preprint arXiv:1611.00068}.

\bibitem[{Stolcke et~al.(1998)Stolcke, Shriberg, Bates, Ostendorf, Hakkani,
  Plauche, Tur, and Lu}]{lm2}
Andreas Stolcke, Elizabeth Shriberg, Rebecca Bates, Mari Ostendorf, Dilek
  Hakkani, Madelaine Plauche, Gokhan Tur, and Yu~Lu. 1998.
\newblock Automatic detection of sentence boundaries and disfluencies based on
  recognized words.
\newblock In \emph{Fifth International Conference on Spoken Language
  Processing}.

\bibitem[{Tilk and Alum{\"a}e(2015)}]{ottokar2015lstm}
Ottokar Tilk and Tanel Alum{\"a}e. 2015.
\newblock Lstm for punctuation restoration in speech transcripts.
\newblock In \emph{Sixteenth annual conference of the international speech
  communication association}.

\bibitem[{Tilk and Alum{\"a}e(2016)}]{ottokar2016blstm}
Ottokar Tilk and Tanel Alum{\"a}e. 2016.
\newblock Bidirectional recurrent neural network with attention mechanism for
  punctuation restoration.
\newblock In \emph{Interspeech}, pages 3047--3051.

\bibitem[{Ueffing et~al.(2013)Ueffing, Bisani, and Vozila}]{CRF1}
Nicola Ueffing, Maximilian Bisani, and Paul Vozila. 2013.
\newblock Improved models for automatic punctuation prediction for spoken and
  written text.
\newblock In \emph{Interspeech}, pages 3097--3101.

\bibitem[{Vaswani et~al.(2017)Vaswani, Shazeer, Parmar, Uszkoreit, Jones,
  Gomez, Kaiser, and Polosukhin}]{attention}
Ashish Vaswani, Noam Shazeer, Niki Parmar, Jakob Uszkoreit, Llion Jones,
  Aidan~N Gomez, {\L}ukasz Kaiser, and Illia Polosukhin. 2017.
\newblock Attention is all you need.
\newblock In \emph{Advances in neural information processing systems}, pages
  5998--6008.

\bibitem[{Yang et~al.(2019)Yang, Dai, Yang, Carbonell, Salakhutdinov, and
  Le}]{xlnet}
Zhilin Yang, Zihang Dai, Yiming Yang, Jaime Carbonell, Russ~R Salakhutdinov,
  and Quoc~V Le. 2019.
\newblock Xlnet: Generalized autoregressive pretraining for language
  understanding.
\newblock In \emph{Advances in neural information processing systems}, pages
  5754--5764.

\bibitem[{Yi and Tao(2019)}]{yi2019self}
Jiangyan Yi and Jianhua Tao. 2019.
\newblock Self-attention based model for punctuation prediction using word and
  speech embeddings.
\newblock In \emph{Proc. ICASSP}, pages 7270--7274.

\end{thebibliography}
\bibliographystyle{acl_natbib}
\end{document}